\documentclass[journal]{IEEEtran}

\usepackage{graphicx}
\usepackage{amssymb}
\usepackage{color}
\usepackage{colortbl}
\usepackage{enumitem}
\usepackage{amsmath}
\usepackage{framed}
\usepackage{subcaption}
\usepackage{url}
\usepackage{float}

\newcommand{\qu}[1]{``#1''}

\newcommand{\beqn}{\begin{eqnarray*}}
\newcommand{\eeqn}{\end{eqnarray*}}
\newcommand{\bneqn}{\begin{eqnarray}}
\newcommand{\eneqn}{\end{eqnarray}}

\newcommand{\yhat}{\hat{y}} 
\newcommand{\parens}[1]{\left(#1\right)}
\newcommand{\prob}[1]{\mathbb{P}\parens{#1}}

\definecolor{gray}{rgb}{0.9,0.9,0.9}
\definecolor{blue}{rgb}{0.498,0.525,0.933}

\ifCLASSOPTIONcompsoc
  \usepackage[nocompress]{cite}
\else
  \usepackage{cite}
\fi

\def\BibTeX{{\rm B\kern-.05em{\sc i\kern-.025em b}\kern-.08em
    T\kern-.1667em\lower.7ex\hbox{E}\kern-.125emX}}

\begin{document}

\markboth{Journal,~Vol.~x, No.~x, x~xxxx}%
{Wu and Kapelner: Predicting Contextual Informativeness}

\title{Predicting Contextual Informativeness for Vocabulary Learning using Deep Learning}

\author{
Tao Wu and~Adam~Kapelner

\thanks{Manuscript xx xx, xxxx; revised xx xx, xxxx. Wu and Kapelner contributed equally to the conceptualization; Wu wrote the code; Wu and Kapelner contributed equally to the drafting and editing of the manuscript.}
}
\IEEEtitleabstractindextext{%
\begin{abstract}
We describe a modern deep learning system that automatically identifies informative contextual examples (\qu{contexts}) for first language vocabulary instruction for high school student. Our paper compares three modeling approaches: (i) an unsupervised similarity-based strategy using MPNet's uniformly contextualized embeddings, (ii) a supervised framework built on instruction-aware, fine-tuned Qwen3 embeddings with a nonlinear regression head and (iii) model (ii) plus handcrafted context features. We introduce a novel metric called the Retention Competency Curve to visualize trade-offs between the discarded proportion of good contexts and the \qu{good-to-bad} contexts ratio providing a compact, unified lens on model performance. Model (iii) delivers the most dramatic gains with performance of a good-to-bad ratio of 440 all while only throwing out 70\% of the good contexts. In summary, we demonstrate that a modern embedding model on neural network architecture, when guided by human supervision, results in a low-cost large supply of near-perfect contexts for teaching vocabulary for a variety of target words.

\end{abstract} 

}

\maketitle
\IEEEdisplaynontitleabstractindextext
\IEEEpeerreviewmaketitle

\ifCLASSOPTIONcompsoc
    \IEEEraisesectionheading{
    \section{Introduction}\label{sec:intro}}
\else
    \section{Introduction}
    \label{sec:introduction}
\fi


\IEEEPARstart{V}{ocabulary} knowledge is foundational for language understanding and production. But vocabulary growth is inherently haphazard: one encounters words naturally in context and these contexts give clues to the word's meaning and these clues aggregate to higher levels of understanding and sophistication \cite{Beck1983, Frishkoff2008}. But not all encounters with the word in context are created equally; some encounters are more \emph{contextually informative} than others. Categorizing context quality at scale is important as it can be used to design effective vocabulary instruction. 

This work extends our 2018 work \cite{Kapelner2018} which hereon will be referred to as our \qu{previous work}. The introduction section therein motivates the importance of modeling contextual informativeness of textual passages (synonymously referred to as a \qu{snippet} or \qu{context}) and has an extensive review of the literature on vocabulary learning and efforts up to that time to model contextual informativeness \cite{Brown2004, Nash2005, Mostow2011, Hassan2011}. The previous work's prediction model was built via supervised learning from a database of human-rated contexts. We then handcrafted features of the snippets to create tabular data for use in classical machine learning. Then we used the Random Forest (RF) algorithm \cite{breiman2001random} to construct a nonlinear generalized model of these features.

The strategy described above was state of the art in 2018. At that time, deep learning architectures with text embedding models had yet to achieve mainstream adoption. Today, deep learning wit text embedding offers a step change in predictive power without the need to engineer features by hand which inevitably will be imperfect, missing important relationships. The processing of language data into numerical representations dates back to the mid-20th century \cite {Arsanjani2023}. Modern neural networks enable the automatic and dynamic encoding of nuanced word meanings in specific contexts. In our setting, this encoding has information about the contextual informativeness of the target word which is what we hope to exploit.



Since 2018, Liu et al. \cite{liu2018content} focused on the domain of children's stories. But Nam et al. \cite{Nam2024} and Valentini et al. \cite{valentini2025} most directly built upon our previous work using deep learning via attention weights. Both these latter teams employed pretrained transformer models, including BERT \cite{Devlinetal2018} and its improved model RoBERTa \cite{liu2019roberta} as well as the non-pretrained models Elmo \cite{peters2018deep} and Gemini \cite{geminiteam2023gemini} to attain impressive predictive performance under both custom and standard evaluation metrics. 

In this paper, we build off our previous work, Nam et al. and Valentini et al. using some of the latest advances in deep learning similar. To make these advancements, we investigate multiple modeling strategies. First is an unsupervised similarity-based modeling strategy. Second is supervised learning approach using the human-rated data from our previous work. This supervised learning begins with a top-of-the-line transformer-based embedding models fine-tuned on unstructured (raw texts) and post-fit on the human-rated data. Third is the previous strategy plus the handcrafted features from the previous work merged together into a larger neural network.



These models are then evaluated under novel metrics we propose to provide a compact and direct lens on their performance in a putative vocabulary-learning system. We find that deep learning for this problem is greatly aided by supervised data and further by including handcrafted features. Our final result is a generalizable system well-suited for teasing out context-word relationships, regardless of word difficulty. 


\section{Methods}\label{sec:met}


We first define our modeling goal: we seek a binary classifier $g$ that takes two inputs (1) a target word we wish to teach and (2) a short textual snippet (context) that contains the target word and outputs either \emph{use} in a vocabulary-teaching system or \emph{not use} in a vocabulary-teaching system. 


\subsection{Raw Data}\label{subsec:raw_data}

Our database of short textual snippets was the original DictionarySquared database \cite{adlof2019accelerating} comprised of 42-65 word snippets culled between 2008-2009 using the now defunct Google web search API. Each snippet contains the target word (in its exact inflection form) at least once, nearly all snippets begin at the beginning of a sentence, nearly all snippets end at the end of a sentence, nearly all snippets do not contain a paragraph break (and coming from within one HTML tag) and all snippets were devoid of non-English non-numeric characters. As the goal is to build a system for teaching high-school level English vocabulary, we use Adlof et al.'s \cite{adlof2019accelerating} categorization of each unique target word into 10 difficulty \qu{bands} based on works by vocabulary experts (see previous work's Section 2.1.1). For each band, we selected $\approx 100$ words for a total of 937 words and for each word we selected at least 13 contexts (and most often many more) for a total of 67,807 total contexts (this number differs slightly from our previous work as we eliminated some duplicates).

\subsection{Human Labels}\label{subsec:human_labels}

Each context was hand-labeled using workers on Amazon’s Mechanical Turk (MTurk). Ten unique workers rated each context for a total of $\approx 700,000$ ratings with quality checks described in Section 2.1.2 of the previous work. Each label was one of the following ordinal values where the workers saw these descriptions but not the numeric values:\\

\begin{itemize}
    \item[(+2)] \textbf{Very Helpful} After reading the context, a student will have a very good idea of what this word means,
    \item[(+1)] \textbf{Somewhat Helpful},
    \item[(0)] \textbf{Neutral} The context neither helps nor hinders a student’s understanding of the word’s meaning,
    \item[(-1)] \textbf{Bad} This context is misleading, too difficult, or otherwise inappropriate.\\
\end{itemize}

The final label for each context, which we treat as the gold-standard, is then defined as the simple arithmetic average of the 10 workers' ordinal values and thus $y_i \in [-1,+2]$. The average rating was $0.59 \pm 0.53$ where $\approx 15\%$ of contexts had $y_i < 0$ and  $\approx 19\%$ of contexts had $y_i > 1$. We will term the former as \qu{misdirective} contexts as these are likely to confuse a student about the meaning of the target word and we we will term the latter as \qu{directive} contexts as they are likely to aid a student in learning the target word. 



\subsection{Feature Engineering}\label{subsec:feature_engineering}

We defined and computed 615 separate characteristics of each context. We used google books n-grams features, synonym features, numbers of colocated words features and other natural language processing (NLP) features. As these features are not an integral part of this work, we refer the reader to Section 2.1.3 of our previous work for details. We now turn to a description of our models.

\subsection{The Unsupervised Model}\label{subsec:deep_unsup}

We hypothesize that the proximity between vectors in the high-dimensional embedding space reflects how much semantic information a context possesses about a word. This proximity will provide a useful proxy measure for our purposes. We define proximity as the cosine similarity $\in [-1,1]$ between a target word's $w$ vector $\vec{v}_{w}$ (and its surrounding context's $c$ vector $\vec{v}_{c}$ in embedding space. We explain how we use this cosine similarity metric to produce the binary classifier $g$ that returns \emph{use} or \emph{not use} in Section~\ref{sec:res}.

To compute cosine similarity, we employed two strategies: the Masked and Permuted Pre-training  for Language Understanding (MPNET) \cite{song2020mpnet} and Qwen3 \cite{qwen3}.

\subsubsection{MPNET}\label{subsubsec:unsupervised_mpnet}

This is a transformer-based embedding model that improves on earlier ones such as BERT and RoBERTa by integrating permuted language modeling with relative position encoding. Most notably, MPNet has been shown to excel at uniform contextualization, i.e., enriching every token in a sentence sequence with the same depth of contextual understanding, making it well-suited for word–context similarity, in our case, with an underutilized vector-extraction technique. The following are model specifications, settings, and hyperparameters for our implementation of MPNet:
\vspace{0.2cm}
\begin{itemize}
    \item \textbf{Model:} \texttt{all-mpnet-base-v2}, pretrained on a large-scale text corpus of over 160GB
    \item \textbf{Model specification:} 12 transformer encoder layers, hidden size 768, 12 attention heads, $\sim$110M parameters
    \item \textbf{Input format:} Target word and its surrounding passage
    \item \textbf{Pooling strategy:} Mean pooling of token embeddings
    \item \textbf{Informativeness measure:} Cosine similarity between target word embedding and context embedding
    \item \textbf{Batch size:} 64
    \item \textbf{Maximum sequence length:} 512 tokens
\end{itemize}
\vspace{0.2cm}

For each word–context pair, we extract vectors from MPNet’s final hidden layer, which is often thought of as the final matrix slice of a 3D tensor. Since a model-specific tokenizer may sometimes break down a word into multiple tokens before passing them through the model, the target word vector is computed by mean-pooling (averaging) the token vectors aligned with the target word's character span. That is, words and vectors do not always have a 1:1 correspondence such that mapping the characters of a word to its token(s) is required. Meanwhile, the context vector can be easily computed by mean-pooling the vectors of all remaining tokens in the passage. Thus, proximity is defined as
\bneqn\label{eq:cosine_sim}
\text{Proximity}(\text{w}, \text{c}) \;=\;
\operatorname{cos\_sim}\!~\parens{\text{mean}(\vec{v}_{w}),\; \text{mean}(\vec{v}_{c})}.
\eneqn
Both mean vectors are normalized before taking their dot product (i.e., their cosine similarity). This formulation measures how semantically integrated a word is within its surrounding context, thus providing a proxy as to how well a context can inform readers about a particular word's meaning without any supervision or fine-tuning on domain-specific data. 

\subsubsection{Qwen3}\label{subsubsec:unsupervised_qwen3} 

This open-source embedding model boasting 4 billion parameters is known to be top-performing for a wide variety of tasks. (The model specifications, settings, and hyperparameters for our implementation is found in the following section). Although Qwen3 uses autoregressive-style embeddings providing a strong contextualization basis, the final end-of-sequence token vector, albeit exceptionally contextualized and most representative of the passage as a whole, does not encode the contextualized meaning of a word in isolation. Such a method of deriving the text's semantics inadvertently de-emphasizes word-context distinction, leading to \emph{lower} similarity between them. This is deleterious in our setting. 

\begin{figure}[ht]
  \centering  \includegraphics[width=2.6in]{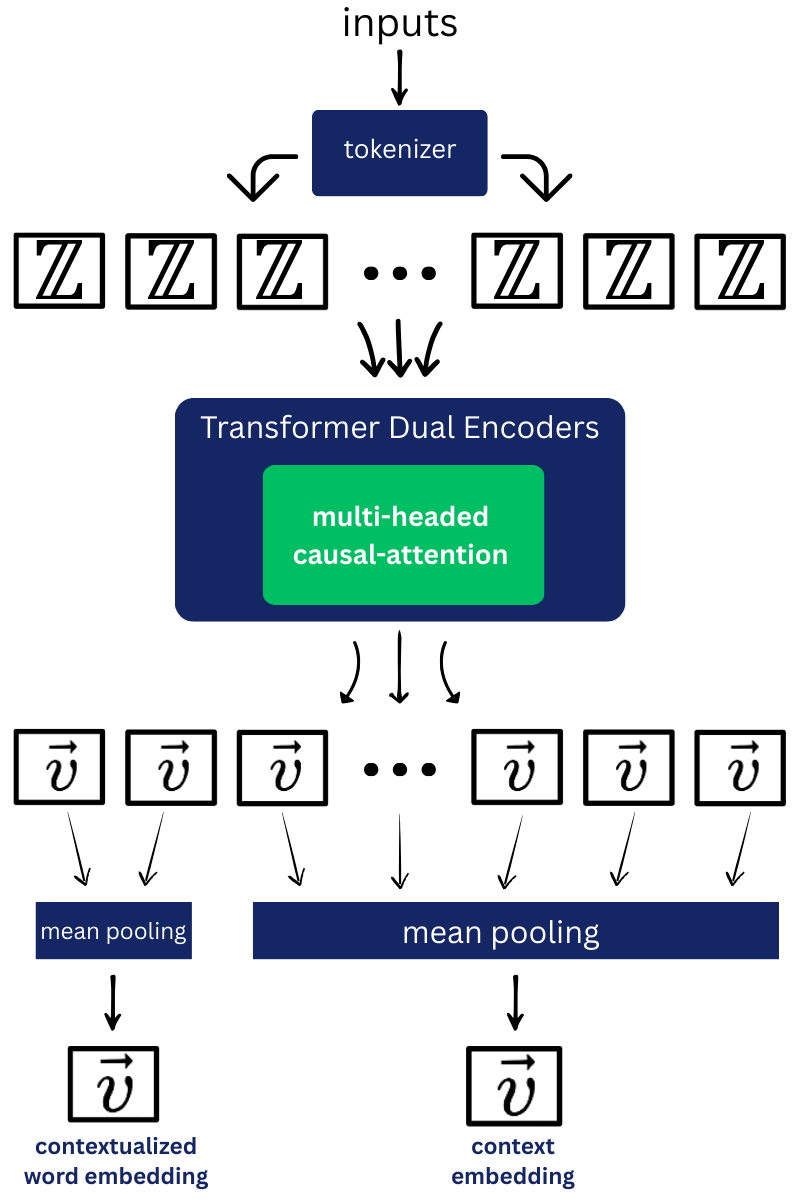}
  \caption{Qwen3 multi‑head causal‑attention mechanism\cite{zhang2025qwen3} followed by mean‑pooling token vectors whose positions corresponding to a target word are arbitrary in this illustration.}
  \label{fig:qwen3_attention}
\end{figure}

We believe we can fix this issue described above by comparing the mean-pooled embedding of a target word against that of the rest in a passage after the multi-headed self-attention mechanism \cite{vaswani2017attention}. This is illustrated in Figure~\ref{fig:qwen3_attention}.

\subsection{The Supervised Learning Model}\label{subsec:deep_sup}

Similarity between target word and context alone, albeit using a generalized well-trained embedding, likely cannot capture what we seek to model because it lacks the targeted objective function defined in the units of a human gold-standard response. We now make use of the human labels from the MTurk workers here and use the $y_i$'s as the prediction target directly in the neural network optimization by installing a nonlinear regression head to the model's output. This model hence produces a prediction $\yhat \in [-1, +2]$. We explain how we use this $\yhat$ prediction to produce the binary classifier $g$ that returns \emph{use} or \emph{not use} in Section~\ref{sec:res}.

Due to insufficient GPU memory in the computing environment, we adopt a smaller Qwen3 Embedding with 0.6B parameters. The following are model specifications, supervised setup stats, and hyperparameters:
\vspace{0.2cm}
\begin{itemize}
    \item \textbf{Model:} \texttt{Qwen/Qwen3-Embedding-0.6B}, pretrained on 36 trillion tokens across 119 languages
    \item \textbf{Model specification:} 24 transformer layers, hidden size 1024, 16 attention heads, $\sim$0.6B parameters
    \item \textbf{Input format:} Instruction-augmented (task prompt + target word + context)
    \item \textbf{Pooling strategy:} End-of-sequence (EOS) token embedding
    \item \textbf{Regression head:} 3-layer MLP with 512 hidden units, ReLU activations, and dropout rate 0.1
    \item \textbf{Loss function:} SmoothL1Loss (Huber)
    \item \textbf{Optimizer:} AdamW (learning rate $1 \times 10^{-5}$, weight decay 0.01)
    \item \textbf{Batch size:} 16
    \item \textbf{Epochs:} 2
    \item \textbf{Maximum sequence length:} 512 tokens
\end{itemize}
\vspace{0.2cm}


Additionally, Qwen3 is instruction-aware \cite{zhang2025qwen3}, meaning that its embeddings can be conditioned on English prompts. To exploit this capability, we experimented with several prompts to find the most effective phrasing of the task description. The \qu{document} was always the plain text of the context. We name these queries and describe them below.

\begin{enumerate}[label=(\roman*),leftmargin=*]
\item \textbf{Plain:} \qu{[target word]}
\item \textbf{Instruction-Augmented:} ``Rate how contextually informative the context is about [target word]''
\item \textbf{Hybrid:} ``What is the definition of [target word]?''
\end{enumerate}


In our experiments (unshown), we found the \textbf{Instruction-Augmented} to yield the best results.

\subsection{The Supervised Learning + Handcrafted Features Model}\label{subsec:deep_sup_plus}

We use the model from Section~\ref{subsec:deep_sup} and add the 615 handcrafted features described in Section~\ref{subsec:feature_engineering} to produce a \qu{hybrid}, \qu{feature-augmented deep learning}, or \qu{ensemble} model as termed in the literature. The integration of the features is handled via late fusion of a tabular layer as follows. We locate the end-of-sequence hidden state from Qwen3, which is a 1024-dimension vector. We then normalize each of the 615 features to have mean zero and standard deviation one and concatenate the 615-dimension vector. The total 1639-dimension vector is fed through three layers: (1) linear with output 512 dimensions and dropout of 0.1 (2) linear with output 512 dimensions and dropout of 0.1 and finally (3) the linear output layer with no activation whose output is scalar. This model is thus both \qu{wide and deep} \cite{cheng2016wide}.


\subsection{Model Prediction and Model Errors}\label{subsec:model_prediction}

The models we described above output a continuous metric $\yhat$ bounded within [-1, +1] for the model described in Section~\ref{subsec:deep_unsup} and within [-1, +2] for the models described in Sections~\ref{subsec:deep_sup}-\ref{subsec:deep_sup_plus} where a higher number denotes higher degree of the context's directive quality. Per our problem setting definition, the $g$ function must use this intermediate continuous output to decision a word-context pair to \textit{use} or \textit{not use}. Thus, we must pick a threshold $\yhat_0$ where if $\yhat > \yhat_0$, our decision will be to \textit{use} this context for this target word. This $\yhat_0$ defines a binary decision with asymmetric costs. If we decision \textit{use} for a misdirective context, the student gets confused; if we decision \textit{not use} for a directive context, we lose a quality context from our pool (which we will term \qu{throwout}). 

Given the large number of contextual examples we can cull from the internet, the cost of throwout is likely much lower than the cost of serving a student a confusing context. (Although for very rare target words, the number of contexts available publicly is truly not infinity, e.g., \qu{pulchritudinous} appears once in 100,000,000 words as measured in Google's 1-trillion-word corpus \cite{brants2006web}). Sliding the threshold from the intermediate continuous output upwards makes the decision more selective hence increasing the proportion of contexts thrown out (which we will term the \qu{throwout rate}) but increases the ratio of directive contexts to misdirective contexts. We will define the good-to-bad ratio as the number of contexts with true $y>1$ divided by the number of contexts with true $y<0$. Thus we have two useful metrics that we trade off over values of $\yhat_0$: the throwout rate and the good-to-bad ratio. It is difficult to combine these two into one holistic scalar cost metric. This is analogous to trading off false positives and false negatives along a receiver-operator curve (ROC) or a detection-error-tradeoff (DET) curve \cite{martin1997det} where the specific point on the curve to implement in the final system remains a contextual decision. Herein, we introduce a similar curve: the \qu{Retention Competency Curve} (RCC) that visualizes the trade-off between the nutritiousness of the curated content (the good-to-bad ratio) and the lost proportion of informative contexts (the throwout rate) across various threshold settings. We choose this new curve for evaluating our performance as the ROC (and the DET) is insufficient for fully capturing performance in our setting, similar to as discussed by Muschelli \cite{muschelli2020roc}. To compare models, we need a metric that measures the whole curve holistically. As common when using the ROC, we will use the \textit{area under the curve} (AUC) estimated through trapezoidal numerical integration \cite{fawcett2006introduction}.


We reiterate that the standard scalar out-of-sample regression metrics (for the supervised models) such as RMSE and $R^2$ are insufficient for assessing the educational utility of our models in this setting as they are at best tangentially related to the holistic cost function. This makes our work difficult to compare with the work of Nam et al. and Valentini et al.

\subsection{Performance Validation}\label{subsec:perf_valid}

As described in Section 1.2 of our previous work, there are two types of validation which make sense in our setting. The models are trained on $\approx$ 1,000 words. Advanced vocabulary consists of 10-20,000 academic technical and low-frequency literary words. If we want to design a system that scales to include any number of such target words, we would like to know how well our models perform on target words that it has \textit{not trained on}. This we will term the [word-unseen] regime. We also would like our models to \qu{beef up} their library of nutritious contexts for target words already trained on and thus we may also wish to know how well our models perform on target words we explicitly trained on. This we will term the [word seen] regime.


Validation details be found in Section 5.2 of our previous work but we summarize here. For the [word-unseen] regime, we adopt a 10-fold cross-validation procedure on the 67807 contexts stratified over the 937 unique target words, grouped by band (the approximate word difficulty metric ranging from 1 to 10). In each fold, we construct a holdout of target words (and their associated contexts) by randomly sampling 10\% of the target words from each band, thus setting up the sample to mirror the proportions of all 10 bands. Although not every holdout set is of equal size as a result of varying numbers of contexts per target word (minimum 13 contexts per word), we believe this protocol ensures that performance honestly reflects the models' generalization across the entire spectrum of vocabulary difficulty rather than exhibiting bias toward the \qu{easier} words. This protocol specifically addresses the predicting behavior observed in our previous work, where the old model tended to perform disproportionately well on the words in lower difficulty bands. For the [word seen] regime, our holdout set construction would be simple: we randomly sample 10\% of the contexts within each target word.

To sum up: validation consists of creating out-of-sample RCCs (see Section~\ref{subsec:model_prediction}) for both the [word unseen] and the [word seen] regimes in all three models (Sections~\ref{subsec:deep_unsup}-\ref{subsec:deep_sup_plus}). Since [word unseen] is arguably the more important regime, we do not present results for the [word seen] regime herein to maintain focus and brevity. We note, however, that model  performance is better in this latter regime, as expected.


\section{Results}\label{sec:res}

All experiments were conducted in Google Colab using either an NVIDIA A100 GPU with 40GB of memory or NVIDIA T4 GPU with 16GB of memory. Model training and inference were implemented in \texttt{Python 3.10} with the \texttt{PyTorch} deep learning framework and the \texttt{transformers} library. Data preprocessing and analysis were performed using \texttt{pandas}, \texttt{NumPy}, and visualization in \texttt{matplotlib}. For baseline reconstruction, we used \texttt{scikit-learn}. 

For each model, we show the performance results of Section~\ref{subsec:model_prediction} as a table. As the threshold was varied over many values, we only show a subset of interest. In each table, we mark the point of 70\% throwout rate to have a single point of comparison. (This is a reasonable throwout rate given that some target words have few contexts). The RCC is generated from columns 5 and 6 (from all rows) which we illustrate in one plot simultaneously for all three models.


Before we describe the performance results of our three models, we first summarize the out-of-sample RF results for the [word unseen] regime from our previous work (see Table 2 therein) to provide baseline performance. At the reference 70\% throwout rate, the good-to-bad ratio is merely 11. The good-to-bad ratio maxes near 70 but at a high cost: the throwout rate is approximately $97\%$. This near 100\% toss rate is too high to be useful in practice as a real vocabulary-teaching system would have rare target words and thus nearly zero contexts accepted.

Before we describe the performance results of the unsupervised model, we discuss intermediate results that may be of interest to the reader. We would like to understand how the cosine similarity of Equation~\ref{eq:cosine_sim} is correlated with human labels. We calculate Pearson's correlation coefficient $r$ and Spearman’s rank correlation coefficient $\rho$ where the latter would likely be a more robust metric and for this reason it is likely it is also the primary metric used in Valentini et al. \cite{valentini2025}. For the MPNet model of Section~\ref{subsubsec:unsupervised_mpnet}, $r = 0.0979$ and $\rho = 0.1046$. For the Qwen3 model of Section~\ref{subsubsec:unsupervised_qwen3}, $r = 0.0652$ and $\rho = 0.0395$. These metrics suggests that both models capture only a small modicum of the contextual informativeness as judged by the average of 10 humans. Qwen3’s autoregressive embedding style underperforms MPNET. Likely, they will will both suffer in the performance metrics of interest when compared to the supervised approaches. 

\begin{figure*}[htp]
\vspace*{0.1cm} 
\centering
\includegraphics[width=7in]{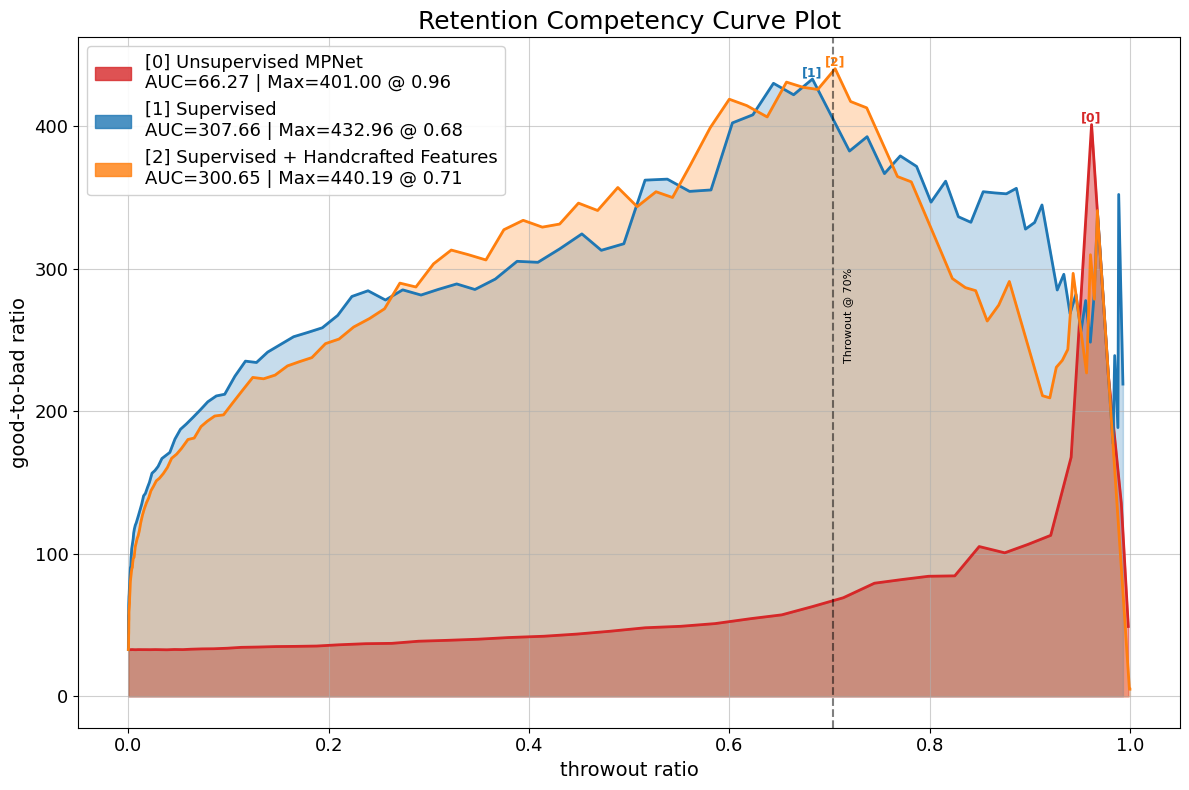}
\caption{RCC plots for all three models with AUC metrics (see legend).}
\label{fig:rcc_auc}
\end{figure*}

We now turn to the results in the performance metrics we find important. Table~\ref{tab:mpnet_sweep} shows a subset of the unsupervised model's performance as a function of the threshold, the cosine similarity of Equation~\ref{eq:cosine_sim}. Table~\ref{tab:qwen3_sweep} shows a subset of the supervised model's performance and Table~\ref{tab:qwen3_hybrid_sweep} shows a subset of the supervised plus handcrafted features model's performance. Figure~\ref{fig:rcc_auc} traces out the RCC's for all three models, displays the computed AUC metrics and marks the 70\% throwout rate reference point.


\begin{table*}[htp]
\centering
\begin{tabular}{ccccccr}
Proximity metric threshold & $\prob{Y < 0}$ & $\prob{Y \in [0,0.5)}$ & $\prob{Y \geq 1}$ & throwout rate & good-to-bad ratio & \# accepted \\
  \hline\hline
0.205 & 0.0141 & 0.1371 & 0.4633 & 0.0000 & 32.7902 & 67807 \\
0.215 & 0.0141 & 0.1371 & 0.4633 & 0.0000 & 32.7902 & 67807 \\
\vdots &&&&&& \\
\rowcolor{yellow} 0.835 & 0.0084 & 0.0847 & 0.5282 & 0.6832 & 62.9937 & 18844 \\
\rowcolor{blue}   0.845 & 0.0077 & 0.0815 & 0.5313 & 0.7138 & 69.1462 & 16920 \\
\rowcolor{yellow} 0.855 & 0.0067 & 0.0786 & 0.5343 & 0.7449 & 79.3465 & 15000 \\
\rowcolor{yellow} 0.865 & 0.0065 & 0.0754 & 0.5364 & 0.7730 & 81.9770 & 13296 \\
\rowcolor{yellow} 0.875 & 0.0064 & 0.0742 & 0.5414 & 0.7989 & 84.2133 & 11666 \\
\rowcolor{yellow} 0.885 & 0.0065 & 0.0693 & 0.5465 & 0.8251 & 84.5077 & 10051 \\
\rowcolor{yellow} 0.895 & 0.0053 & 0.0684 & 0.5539 & 0.8495 & 105.0667 & 8536 \\
\vdots &&&&&& \\
0.985 & 0.0000 & 0.0000 & 0.7500 & 0.9999 & nan & 4 \\
0.995 & nan & nan & nan & 1.0000 & nan & 0 \\
\end{tabular}
\caption{Out of sample results for the unsupervised MPNET model  (Section~\ref{subsubsec:unsupervised_mpnet}) for the [word unseen] regime.}
\label{tab:mpnet_sweep}
\end{table*}

\begin{table*}[htp]
\centering
\begin{tabular}{ccccccr}
$\yhat_0$ threshold & $\prob{Y < 0}$ & $\prob{Y \in [0,0.5)}$ & $\prob{Y \geq 1}$ & throwout rate & good-to-bad ratio & \# accepted \\
  \hline\hline
-0.415 & 0.0141 & 0.1371 & 0.4633 & 0.0000 & 32.7902 & 67807 \\
-0.405 & 0.0141 & 0.1371 & 0.4633 & 0.0000 & 32.7902 & 67807 \\
\vdots &&&&&& \\
\rowcolor{yellow} 0.845 & 0.0026 & 0.0272 & 0.6801 & 0.1936 & 258.4796 & 37248 \\
\rowcolor{yellow} 0.855 & 0.0026 & 0.0261 & 0.6887 & 0.2090 & 267.1828 & 36082 \\
\rowcolor{yellow} 0.865 & 0.0025 & 0.0246 & 0.6973 & 0.2232 & 280.4713 & 34993 \\
\rowcolor{yellow} 0.875 & 0.0025 & 0.0233 & 0.7057 & 0.2393 & 284.4643 & 33859 \\
\rowcolor{yellow} 0.885 & 0.0026 & 0.0224 & 0.7138 & 0.2567 & 277.9643 & 32709 \\
\rowcolor{yellow} 0.895 & 0.0025 & 0.0213 & 0.7224 & 0.2739 & 285.1000 & 31573 \\
\vdots &&&&&& \\
\rowcolor{blue}  1.105 & 0.0021 & 0.0069 & 0.8692 & 0.7021 & 406.8696 & 10766 \\
\vdots &&&&&& \\
1.845 & nan & nan & nan & 1.0000 & nan & 0 \\
1.855 & nan & nan & nan & 1.0000 & nan & 0 \\
\end{tabular}
\caption{Out of sample results for the supervised model (Section~\ref{subsec:deep_sup}) for the [word unseen] regime.}
\label{tab:qwen3_sweep}
\end{table*}

\begin{table*}[htp]
\centering
\begin{tabular}{ccccccr}
$\yhat_0$ threshold & $\prob{Y < 0}$ & $\prob{Y \in [0,0.5)}$ & $\prob{Y \geq 1}$ & throwout rate & good-to-bad ratio & \# accepted \\
  \hline\hline
-1.105 & 0.0141 & 0.1371 & 0.4633 & 0.0000 & 32.7902 & 67807 \\
-1.095 & 0.0141 & 0.1371 & 0.4633 & 0.0000 & 32.7902 & 67807 \\
\vdots &&&&&& \\
\rowcolor{yellow} 0.845 & 0.0027 & 0.0309 & 0.6777 & 0.2103 & 250.5859 & 36606 \\
\rowcolor{yellow} 0.855 & 0.0026 & 0.0298 & 0.6846 & 0.2249 & 259.0106 & 35565 \\
\rowcolor{yellow} 0.865 & 0.0026 & 0.0285 & 0.6919 & 0.2408 & 265.0000 & 34472 \\
\rowcolor{yellow} 0.875 & 0.0026 & 0.0273 & 0.6994 & 0.2558 & 271.8488 & 33428 \\
\rowcolor{yellow} 0.885 & 0.0024 & 0.0259 & 0.7075 & 0.2711 & 289.8228 & 32364 \\
\rowcolor{yellow} 0.895 & 0.0025 & 0.0243 & 0.7156 & 0.2871 & 287.1026 & 31292 \\
\vdots &&&&&& \\
\rowcolor{blue}  1.125 & 0.0020 & 0.0089 & 0.8676 & 0.7057 & 440.1905 & 10655 \\
\vdots &&&&&& \\
1.995 & 0.1364 & 0.0909 & 0.6818 & 0.9995 & 5.0000 & 22 \\
2.005 & nan & nan & nan & 1.0000 & 5.0000 & 0 \\
\end{tabular}
\caption{Out of sample results for the supervised + handcrafted features model (Section~\ref{subsec:deep_sup_plus}) for the [word unseen] regime.}
\label{tab:qwen3_hybrid_sweep}
\end{table*}

The unsupervised model is greatly outperformed by the supervised models, a result we expected given its lack of a targeted objective function. At the reference point, the unsupervised model achieves a good-to-bad ratio of 69 while the supervised models achieve a ratio of 407 and 440. The unsupervised model has an AUC of its RCC of 66 compared to an AUC of the RCC of 308 and 301 for the supervised models. However, the unsupervised model still outperforms the previous work's RF model by a factor of 6:1 which illustrates the power of MPNET's pretraining on a large dataset.

The two supervised models perform similarly as measured by the AUC metric. However, the good-to-bad ratio is better for the supervised + handcrafted features model as measured by the good-to-bad ratio at the reference point and the maximal point. Thus, we would conclude it is a better model to be used for implementation in a real vocabulary-teaching system. 

The supervised models outperform the previous work's RF model by an astounding 37-40x at the reference point. Measured another way, the previous work's RF good-to-bad ratio at $94.5\%$ throwout rate attains a good-to-bad ratio of 125:1 --- an output that our deep learning supervised model produces at only a $1\%$ throwout rate. Thus, the supervised deep learning models here operate at almost 1\% of the cost of the previous work's model. Deep learning is fully superior to tabular machine learning on handcrafted features.

We now report classic regression metrics. As described in Section~\ref{subsec:model_prediction}, classic regression metrics are insufficient metrics to describe performance for our modeling setting. However, to provide comparison with previous works \cite{Nam2024, valentini2025} (even though our work differs substantially in its setup and focus), we provide the results for our supervised model of Section~\ref{subsec:deep_sup} and our supervised plus handcrafted features model of Section~\ref{subsec:deep_sup_plus}. We also introduce a \qu{null model} which merely predicts the sample average of the gold-standard labels ($\approx 0.94$). This null model provides a sanity check on our modeling efforts. We also include results of the previous work's RF model. All results are found in Table~\ref{tab:standard_metrics}. 

\begin{table}[H]
\centering
\setlength{\tabcolsep}{6pt}
\begin{tabular}{lcc}
\rowcolor{gray}\textbf{Model} & \textbf{RMSE} & $\mathbf{R^2}$ \\ \hline
Null model                     & 0.40 & 0 \\
RF (from our previous work)          & 0.36 & 0.18 \\
Supervised  & 0.31 & 0.38 \\
Supervised + Handcrafted Features         & 0.30 & 0.42 \\
\end{tabular}
\caption{Out of sample regression metrics comparison for supervised models herein and in our previous work for the [word unseen] regime.}
\label{tab:standard_metrics}
\end{table}

As noted previously, the deep learning architectures present a large accuracy improvement from the previous work. Additionally, the supervised plus handcrafted features model outperforms the supervised without the handcrafted features model in accuracy although the latter has a slightly lower AUC (see Figure~\ref{fig:rcc_auc}). Our RMSE figures are in line with previous works as seen from Table 2 in Valentini et al. \cite{valentini2025}.


\section{Discussion}\label{sec:discussion}

We used modern deep learning models to construct a scalable, generalizable system to produce high-quality curation of contexts to aid students' vocabulary learning. We considered three models: unsupervised, supervised and supervised plus the handcrafted features of our previous work. We also introduced novel evaluation metrics which were necessary due to the asymmetric costs of discarding bad contexts versus accepting good ones. The three deep learning models comfortably outperform our previous work's efforts to model context quality using handcrafted features and standard tabular machine learning. This is expected, as (1) deep learning architectures leverage vast semantic knowledge captured within their pre-trained representations, (2) the architecture employs automated feature extraction and (3) multilayer perceptrons act as universal approximators capable of modeling complex, non-linear functional relationships. However, we were not expecting the gains to be so large. Pinning the throwout rate to 70\%, the unsupervised model, which did not have the luxury of even seeing our response metric, performs $6\times$ better in the good-to-bad context ratio than the previous work's model. The supervised model performs $407\times$ better and the supervised model with the handcrafted features performs $440\times$ better.

Although the embedding space on modern architectures such as Qwen3 is broad, there still seems to be marginal gains for adding handcrafted features depending on how you assess performance. Some of these handcrafted features such as part-of-speech, are likely redundant and thus cause overfitting, reducing performing. However, there are some features such as n-gram frequencies on large corpuses that can increase performance as Qwen3 or MPNet might not be able to represent n-gram frequencies as they only are trained on local context windows and thus, likely could not have internalized global statistics of the English language. Thus our results echo others who find that hybrid feature-augmented modeling shows performance gains in linguistic modeling settings \cite{shah2020fusion, zaharia2021, faseeh2024hybrid, Nam2024}. However, we only measure marginal gains and thus the labor of computing these features in a production system is likely not worth the trouble.

\subsection{Future Directions}\label{subsec:future_direction}


We outline a few future directions roughly in order of complexity.

\subsubsection{Scaling up size}

Larger, more complex models perform better \cite{kaplan2020scaling}. We would like to investigate the larger Qwen3 embedding models with 8B parameters. We expect to demonstrate even more impressive gains.

\subsubsection{Improving the handcrafted feature integration}

Although the gains are marginal herein, other studies have found more impressive gains so investigating other fusion strategies may prove fruitful. Here are some strategies listed in order of complexity. The first is \qu{gated fusion} where instead of raw concatenation, we add a learned gate layer to control how much each modality contributes letting the model learn when to trust the text embedding versus the handcrafted features. We can also try \qu{cross-attention fusion} meaning we treat the handcrafted features as a sequence of 615 tokens, then use cross-attention between the transformer's hidden states and the tabular tokens. This lets the model learn fine-grained interactions. Third, we can try \qu{feature injection} which injects the handcrafted features into the transformer itself (e.g., by adding them as a bias or residual at one or more intermediate transformer layers). Fourth, we can use a separate tabular encoder (e.g., by shrinking the 615 dimensions to 256 dimensions and then concatenating like we described in Section~\ref{subsec:deep_sup_plus}). This would allow for the model to learn its own nonlinear feature interactions before merging and reduces its dimensionality relative to Qwen's end of sequence dimensionality. Lastly, we can use a \qu{FT-Transformer} where each of the handcrafted features gets its own learned embedding, then self-attention is applied across features. The output token from this is then combined with  Qwen's end of sequence vector as described in Section~\ref{subsec:deep_sup_plus}.



\subsubsection{Generalizing to different modalities} 

\begin{figure}[h]
    \centering
    \includegraphics[width=2in]{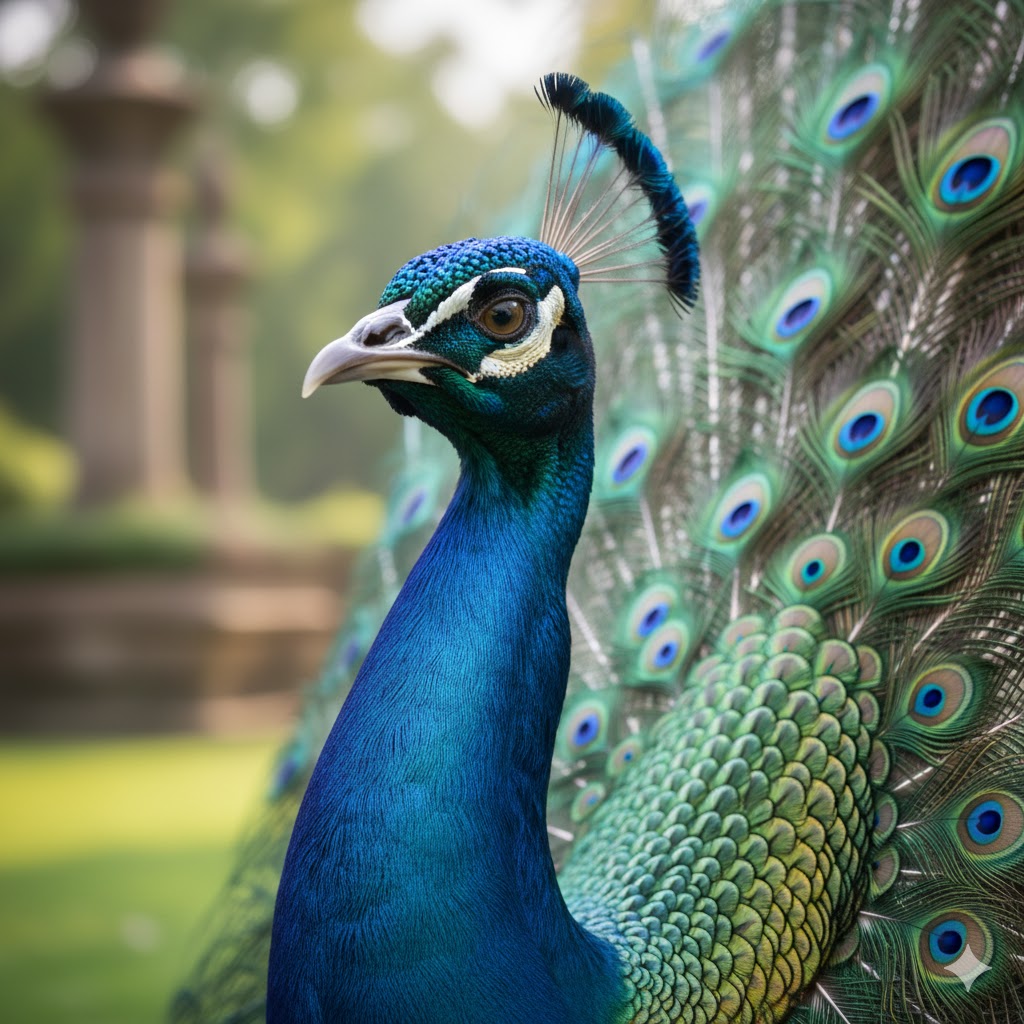}
    \caption{An example image in which a student can learn the target word \textit{supercilious} along with an associated caption such as: \qu{Argyle held his crested head with such localized gravity that one might suspect the rest of the garden was simply revolving around him. As the common pheasants scurried for fallen seed, Argyle offered them nothing but the heavy-lidded appraisal of a bored monarch, his \textit{supercilious} gaze suggesting that their very existence was a mere clerical error in mother nature's ledger.}}
    \label{fig:supercilious}
\end{figure}

We believe our system can generalize from target word to images and videos with audio and transcripts (as implemented in the original DictionarySquared system), expanding the modality and broadening pedagogical efficiency. High-quality royalty-free visual assets can be synthesized using models like Nano Banana \cite{geminiteam2023gemini} or sourced from existing media via fair use, which permits the inclusion of short clips from movies and television. One can learn the target word \qu{supercilious} from reading contexts or you can see it visually with a caption such as in Figure~\ref{fig:supercilious}.


\section{Acknowledgement}\label{sec:acknowledgement}

This research was supported by Grant No 2018112 from the United States-Israel Binational Science Foundation (BSF).

\ifCLASSOPTIONcaptionsoff
  \newpage
\fi
\bibliographystyle{IEEEtran}
\bibliography{ml}

\vspace{-1cm}
\begin{IEEEbiography}[{\includegraphics[width=1in,height=1.25in,clip,keepaspectratio]{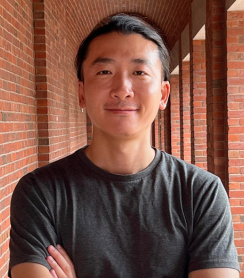}}]{Tao Wu}
earned his B.A. in Mathematics with a minor in Computer Science from Queens College, CUNY and is currently pursuing an M.A. in Statistics at Columbia University. His research interests center on machine learning and causal inference, with a particular focus on developing and applying modern algorithms to uncover fundamental patterns and support prediction-based decision making.
\end{IEEEbiography}

\vspace{-1cm}
\begin{IEEEbiography}[{\includegraphics[width=1in,height=1.25in,clip,keepaspectratio]{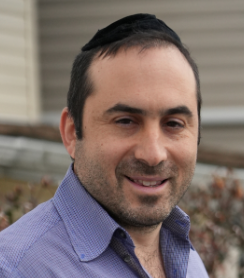}}]{Adam Kapelner}
received his Ph.D. and A.M. in Statistics from the Wharton School of the University of Pennsylvania and his B.S. in Computational Science from Stanford University. He is an Associate Professor of Mathematics at Queens College, CUNY where he heads the undergraduate Data Science \& Statistics program. His research interests include improving randomization designs to improve power in experimental design, machine learning and running social science experiments.
\end{IEEEbiography}





\end{document}